%% file: main.tex
\documentclass[a4paper, 10pt, conference]{ieeeconf}      

\IEEEoverridecommandlockouts                              
\usepackage{url}
\usepackage{amsmath}
\usepackage{amssymb}
\usepackage{algorithm}
\usepackage{algpseudocode}
\usepackage{tikz}
\usepackage{tikz-3dplot}
\usepackage{subcaption}
\usepackage{pgfplots} 
\usepackage{pgfgantt}
\usepackage{pdflscape}
\pgfplotsset{compat=newest} 
\pgfplotsset{plot coordinates/math parser=false}
\usepackage{todonotes}
\usepackage{placeins}
\usepackage{svg}

\newlength\fwidth

\emergencystretch=\maxdimen
\begin{document}

\title{Detection of Endangered Deer Species Using UAV Imagery:\\A Comparative Study Between Efficient Deep Learning Approaches}

\author{
    Agustin Roca$^{1}$, Gastón Castro$^{1,2}$, Gabriel Torre$^{1}$, \\Leonardo J. Colombo $^{3}$, Ignacio Mas$^{1,2}$, Javier Pereira$^{2,4}$, Juan I. Giribet$^{1,2}$
\thanks{${}^1$Artificial Intelligence and Robotics Laboratory (LINAR), Universidad de San Andr\'es, Vito Dumas 284, B1644BID Victoria, Provincia de Buenos Aires, Argentina, {\tt\small aroca@udesa.edu.ar, gcastro@udesa.edu.ar, jgiribet@udesa.edu.ar, torreg@udesa.edu.ar, imas@udesa.edu.ar}}
\thanks{${}^2$Consejo Nacional de Investigaciones Cient\'ificas y T\'ecnicas (CONICET), Argentina.}
\thanks{$^3$Centre for Automation and Robotics (CSIC-UPM), Ctra. M300 Campo Real, Km 0,200, Arganda del Rey - 28500 Madrid, Spain, {\tt\small leonardo.colombo@csic.es}}
\thanks{${}^4$Museo Argentino Bernardino Rivadavia, Av. Angel Gallardo 470, C1405DJR, Capital Federal, Argentina, {\tt\small jpereira@conicet.gov.ar}}
\thanks{The authors were partially supported by  PICT-2019-2371 and PICT-2019-0373 projects from Agencia Nacional de Investigaciones Cient\'ificas y Tecnol\'ogicas, UBACyT-0421BA project from the Universidad de Buenos Aires (UBA), Argentina, from the Spanish Ministry of Science and Innovation, under grants  PID2022-137909NB-C21 funded by MCIN/AEI\-/10.13039\-/501100011033 and the LINC Global project from CSIC ``Wildlife Monitoring Bots'' INCGL20022.}}

\maketitle

\vspace{-1em}
\noindent
\begin{center}
    \parbox{0.95\linewidth}{
        \footnotesize
        © 2025 IEEE. Personal use of this material is permitted.  Permission from IEEE must be obtained for all other uses, in any current or future media, including reprinting/republishing this material for advertising or promotional purposes, creating new collective works, for resale or redistribution to servers or lists, or reuse of any copyrighted component of this work in other works.
    }
\end{center}
\vspace{0.5em}

\thispagestyle{empty}
\pagestyle{empty}

\begin{abstract}
This study compares the performance of state-of-the-art neural networks including variants of the YOLOv11 and RT-DETR models for detecting marsh deer in UAV imagery, in scenarios where specimens occupy a very small portion of the image and are occluded by vegetation. We extend previous analysis adding precise segmentation masks for our datasets enabling a fine-grained training of a YOLO model with a segmentation head included. Experimental results show the effectiveness of incorporating the segmentation head achieving superior detection performance. This work contributes valuable insights for improving UAV-based wildlife monitoring and conservation strategies through scalable and accurate AI-driven detection systems.
\end{abstract}


\section{Introduction}

Monitoring wildlife is crucial for understanding and preserving ecosystems. Traditional observation methods, however, are often labor-intensive, costly, and constrained in coverage \cite{buckland2004advanced, groom2013remote}. The integration of Unmanned Aerial Vehicles (UAVs) offers a transformative approach, enabling extensive area surveillance and remote data collection. Despite these advantages, challenges persist in effectively detecting and identifying species, particularly those that are small or well-camouflaged, such as deer, in aerial imagery. This work examines the application of artificial intelligence techniques to automate the identification of deer in aerial images, addressing a significant challenge in wildlife conservation.

Recent efforts have focused on leveraging UAV-based monitoring for the protection of endangered deer species in diverse ecological settings. These initiatives involve multidisciplinary teams comprising scientists, engineers, and conservationists, working towards improving population monitoring and habitat assessment. A key challenge in these efforts is the automatic detection of deer using aerial imagery, which demands robust image processing and machine learning techniques. UAV surveys have been conducted in various regions, collecting high-resolution imagery to support conservation strategies. The implementation of AI-based detection systems is a crucial step in automating the identification process, reducing human effort, and enhancing monitoring accuracy.

One such initiative is the Pantano Project \cite{ppantano}, dedicated to the conservation of marsh deer (\textit{Blastocerus dichotomus}) in the Paran\'a Delta, Buenos Aires, Argentina. This project, involving scientists from research centers and universities, aims to study the ecology and behavior of marsh deer and develop effective conservation strategies. UAVs have played a critical role in this initiative by enabling remote detection of marsh deer in their natural habitat. Aerial surveys conducted as part of the project have facilitated extensive data collection, which has been instrumental in training AI-based detection models. Collaboration with researchers and trained volunteers has further enhanced the effectiveness of these efforts.

\begin{figure*}[h!]
\centering
\includegraphics[width=0.9\linewidth]{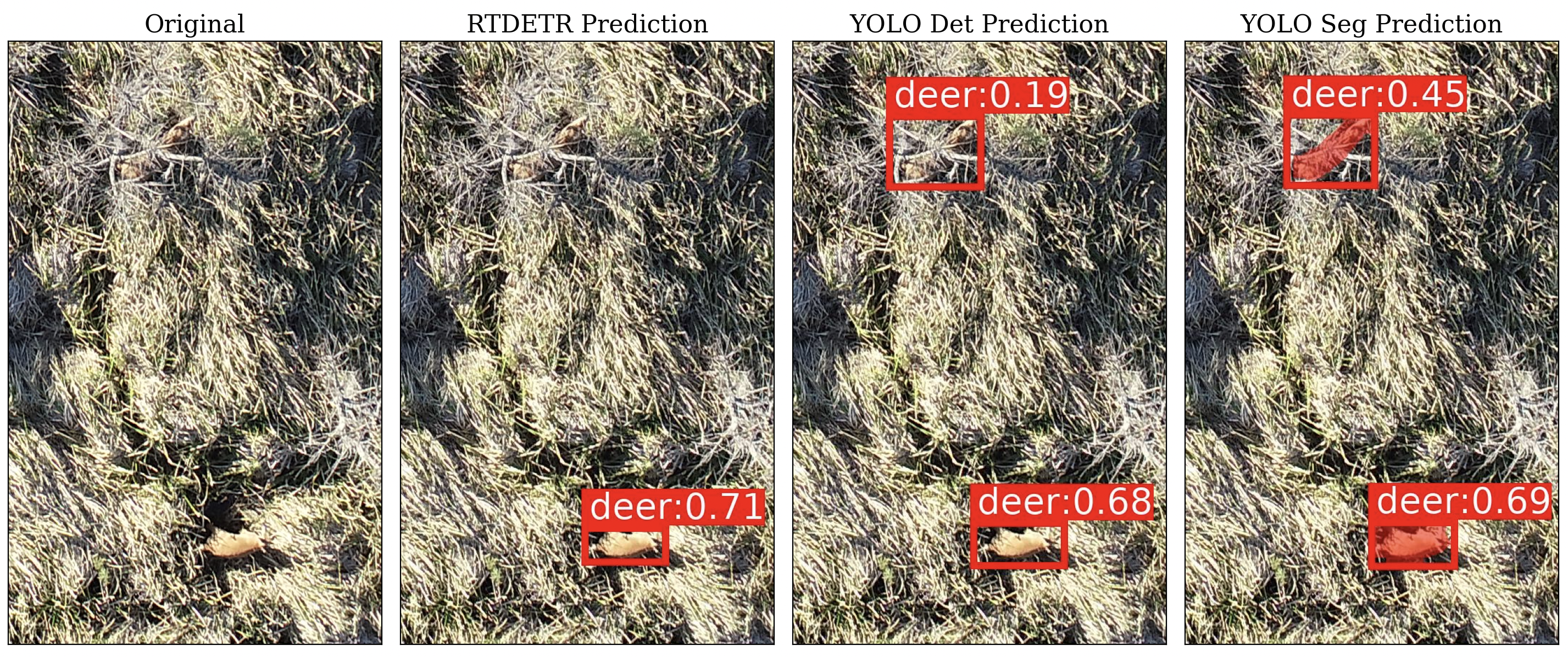}
\caption{Visual comparison of predictions for a sample image from the test set. It shows the predicted class and the confidence score of each detection}
\label{fig:comparison}
\end{figure*}

This study presents the initial results of developing an automated detection system for deer species using deep learning algorithms. Utilizing datasets generated from aerial surveys, object detection models based on YOLO (You Only Look Once) \cite{redmon2016yolo, yolo_ultralytics} and RT-DETR (Real Time DEtection TRansformer) \cite{zhao2024rtdetr} have been trained and evaluated. Some detections using these trained models can be seen in Figure \ref{fig:comparison}. The system's performance is assessed by comparing its detections against expert-verified census data, providing insights into its reliability and potential for large-scale deployment.


\section{Related Work}\label{Sec2}

Over the past decade, UAVs have gained traction as a tool for wildlife observation. Linchant et al. \cite{Linchant2015} provide an extensive review of advancements and challenges in this field up to 2015. Their study highlights key technical limitations, including restricted coverage of large geographic areas, emphasizing the need for improved sampling protocols, inventory methods, and statistical analyses. Additionally, they underscore the potential of combining thermal sensors with high-resolution optical imaging to enhance species differentiation. Another major challenge identified is the lack of regulatory frameworks supporting UAV-based wildlife monitoring.

Since 2015, substantial progress has been made in both regulatory aspects—albeit at a slower pace—and in refining methodologies for data collection and analysis \cite{fust2023increasing, pereira2023}. Advances in sensor fusion techniques have further improved species detection, while artificial intelligence-driven automatic detection strategies have emerged as valuable tools for streamlining species identification and classification.

A variety of approaches have been explored for detecting wildlife in aerial imagery, including object detection algorithms, semantic segmentation techniques, and deep learning-based methods \cite{chabot2016computer, li2019use, sundaram2020fsscaps}. For instance, Barbedo et al. \cite{barbedo2019study} leveraged Convolutional Neural Networks (CNNs) to monitor cattle herds using UAV-acquired imagery. In another study, Brown et al. \cite{brown2022automated} utilized high-resolution UAV images to establish precise ground-truth data, implementing YOLOv5-based object detection for livestock localization and counting in Australian farms. Deep learning models, particularly YOLO, have demonstrated significant potential for UAV-based wildlife detection tasks.

The effectiveness of deep learning in UAV wildlife monitoring is highly dependent on access to large-scale datasets for training algorithms \cite{weiss2016survey}. However, the scarcity of extensive aerial wildlife image datasets has posed a challenge, often requiring adaptation of object detection models originally designed for natural scene images. This adaptation, though widely used, is not always optimal for wildlife detection applications \cite{zheng2021self}. Consequently, UAV-based detection algorithms continue to struggle with issues such as low accuracy, limited robustness, and suboptimal real-world performance \cite{okafor2017operational, kellenberger2018best}.

Several studies have sought to address these limitations. For example, Mou et al. \cite{Mou2023} introduced a real-time species detection model based on YOLOv7, optimized for detecting small targets in UAV-based monitoring. Their work also presented the Wildlife Aerial Images from Drone (WAID) dataset, consisting of 14,375 high-quality UAV images across multiple environmental conditions, featuring six different wildlife species and diverse habitat types. Comparative experiments demonstrated that the SE-YOLO model outperforms others, achieving a mean average precision ($mAP$) of 0.983, with a particularly high $mAP$ of 0.926 for animals occupying fewer than 50 pixels in an image. Similarly, Lyu et al. \cite{Lyu2024} explored RGB and thermal image fusion for deer detection, showing that transferring a model trained on visible spectrum images to thermal imaging is challenging due to the limited detail in thermal data. However, with an appropriately designed neural network, detection performance significantly improves.

\begin{figure*}[h!]
\centering
\includegraphics[width=\linewidth]{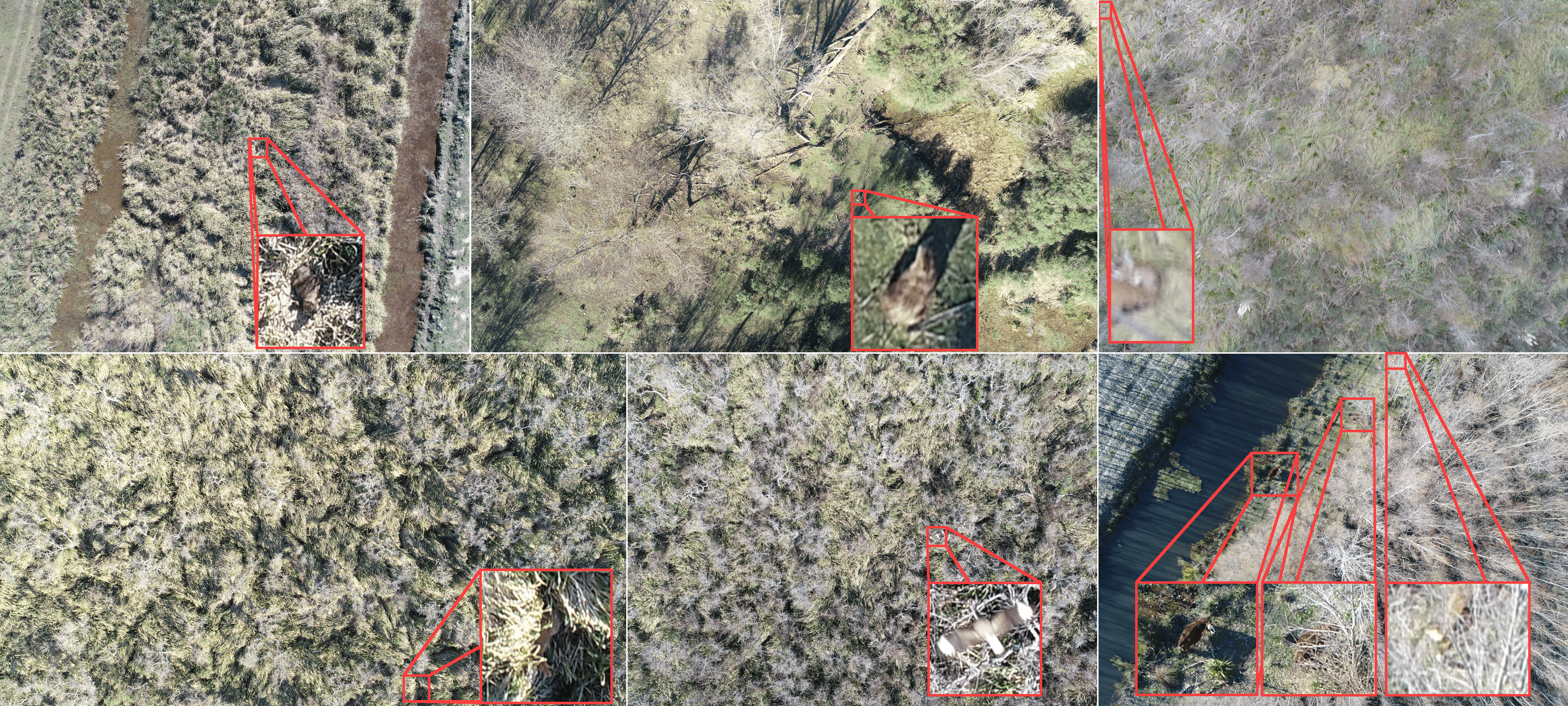}
\caption{Images with different animals spotted during UAV test flights, including capybaras, deer, cows, and birds.}
\label{fig:animales}
\end{figure*}

In a previous related study \cite{Roca2024}, the present authors proposed an approach for the automatic detection of endangered deer species using UAV aerial images combined with the YOLO framework. The model, trained on high-resolution RGB images manually annotated (the same dataset used in this work), achieved an $mAP@10$ of 0.950 in real-world conservation applications. However, when tested with the Pampas deer, a different cervid species, the model exhibited a significant performance drop, highlighting the need for species-specific training data. The current work extends the analysis by performing a comparative evaluation of two state-of-the-art models for object detection tasks, YOLOv11, and RT-DETR. The decision of comparing specifically these two networks is because RT-DETR architecture relies on transformers while YOLOv11 architecture does not. Our dataset is enriched with precise segmentation masks, a novel contribution to wildlife monitoring specific datasets, which further supports more accurate and reliable models training.

In recent years, the Artificial Neural Networks community has witnessed significant advancements with the introduction of Transformer-based architectures, notably the DEtection TRansformer (DETR) \cite{misra2021end} has impacted the object detection field. The DETR architecture starts with a CNN backbone that extracts high-level features from the image. Typically, the ResNet-50 or ResNet-101 \cite{he2016resnet} are used as backbones, generating feature maps that serve as input to Transformer blocks. The self-attention mechanism of the Transformer allows the model to capture long-range correlations between visual features present in the image. Traditional CNN-based detectors struggle with capturing distant correlations due to their localized receptive fields. In contrast, DETR’s Transformer-based decoder attends to all spatial features simultaneously. Moreover, the end-to-end DETR architecture introduces a novel methodology that eliminates the need of the Non-Maximum Suppression (NMS) process required in YOLO models, simplifying the detection pipeline. In this line of work, the Real-Time DETR (RT-DETR) \cite{zhao2024rtdetr} proposes a real-time variant of the DETR architecture showing superior accuracy than the YOLOv8 model in the well-known COCO dataset \cite{lin2014coco}, while also maintaining higher processing speeds.

Building on these advancements, this study compares the performance of two variants of the YOLOv11 model, and the RT-DETR model for detecting marsh deer in UAV-acquired imagery. This comparison is highly valuable, as images captured by UAVs differ significantly from traditional object detection datasets in their characteristics. 
The specimens are represented by a small number of pixels in the images and they are highly occluded by the vegetation. While the primary focus is on marsh deer, the dataset also includes images containing other animals such as cows, capybaras, and birds. This diversity enables the model to demonstrate its capability in handling multiple species within the detection framework. By integrating deep learning techniques with real-world wildlife monitoring, this approach contributes to improving conservation efforts through enhanced UAV-based species identification.

\section{Methodology}\label{Sec3}

\subsection{Data Acquisition and Dataset Characteristics}

A key aspect of utilizing specific tools for automatic wildlife detection is the availability of a comprehensive and diverse dataset to train the algorithms. In this study, data from the Pantano Project \cite{ppantano} was used.

To gather the data for the database within the Pantano Project framework, Phantom 4 Pro UAV quadcopters were employed, each equipped with a 20 MP high-definition camera offering an 84° field of view and an 8.8/24 mm lens, featuring an autofocus system ranging from f/2.8 to f/11 at a 1-meter distance. The cameras were mounted on a three-axis stabilizer. The UAVs flew autonomously, with the flight route uploaded to DJI's proprietary software. Once the UAV was launched, it followed the pre-programmed route, with the ground operator observing remotely.

An initial exploratory study was carried out on May 16 and 17, 2019, during which six flight plans were executed, covering a total of 19.5 km. These flights were designed to evaluate different parameters such as speed, altitude, and image capture timing, aiming to enhance the chances of distinguishing marsh deer from their surroundings across various habitat types in the study area. After the analysis, it was determined that an altitude of 45 m above ground (with a 67.5 m wide transect), a speed of 6.5 m/s, and a photograph taken every 5 seconds—resulting in about 33\% overlap between consecutive images—were optimal. No escape behavior from the deer was recorded during UAV flights with these chosen parameters.

A second set of flights took place on August 6 and 8, 2019, during winter, a time when leaf cover on the poplars and willows in the deer's habitat is reduced. A grid of 1500m (north-south) by 100m (east-west) cells was overlaid on an image of the study area. Transects of 1500m were created through the intersection of north-south and east-west lines. Given the irregular perimeter of the study area, only transects longer than 760m were selected. These transects were numbered and randomly chosen for inclusion in the study until 10\% of the study area was covered. The flight routes were then designed using DJI GS Pro software to cover as many selected transects as possible, within the limits of the UAV’s battery range. Additional transects were incorporated when necessary, particularly when the start and end positions of the initial transects were more than 1500m apart. The first and last 100m of these transects were excluded from analysis to avoid potential double counting of individuals. However, these images were still used for training the algorithms as they provided valuable information.

The flight routes were executed in quick succession, with 500m separating them within a span of no more than 3 hours, to reduce the chance of double counting, as deer might have moved between adjacent transects.

In total, 39,798 photographs were captured across 575 transects during the flights conducted as part of the Pantano Project. Figure \ref{fig:animales} shows some examples of the captured images, where different animals can be observed. These photographs were manually reviewed by a team of four expert scientists and 168 trained volunteers, who cataloged the images containing marsh deer. Each observer analyzed a subset of images following a standardized procedure: digitally zooming in on each image and scanning it left to right and top to bottom for any marsh deer. Sequential overlapping images were also compared to avoid double counting. To ensure accuracy, each photograph was reviewed by at least two independent observers. The final count of marsh deer, excluding multiple detections of the same individual by different observers, was used to estimate the density of marsh deer in the study area.

A total of 231 marsh deer were identified by the experts across the analyzed images. In addition, during the course of work, the network was able to find new unlabeled deer that were later confirmed by experts. The final count is of 285 marsh deer distributed between 262 images. Precise segmentation masks were created for those images isolating pixels corresponding to the specimens. At the moment of writing, we do not know any other wildlife dataset that extracts this type of fine-grained segmentation masks.

\subsection{Models Architecture, Configuration and Training Setup}

\begin{figure*}[h!]
\centering
\def\svgwidth{width=0.8\linewidth}
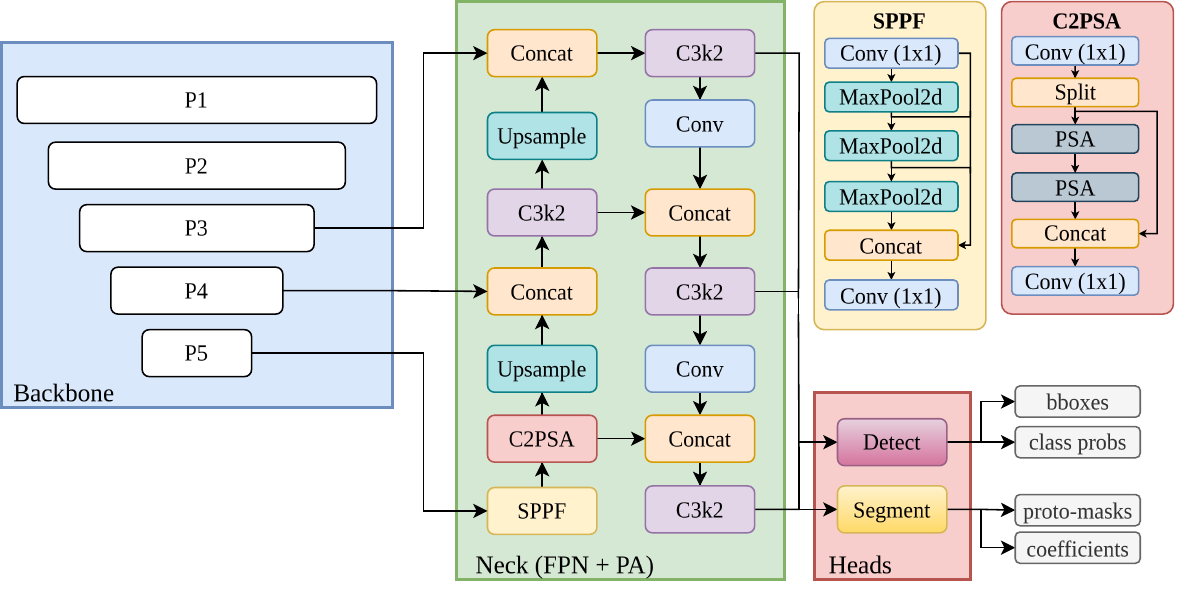
\caption{Diagram of YOLOv11 architecture}
\label{fig:yolov11}
\end{figure*}

\begin{figure*}[h!]
\centering
\def\svgwidth{width=0.8\linewidth}
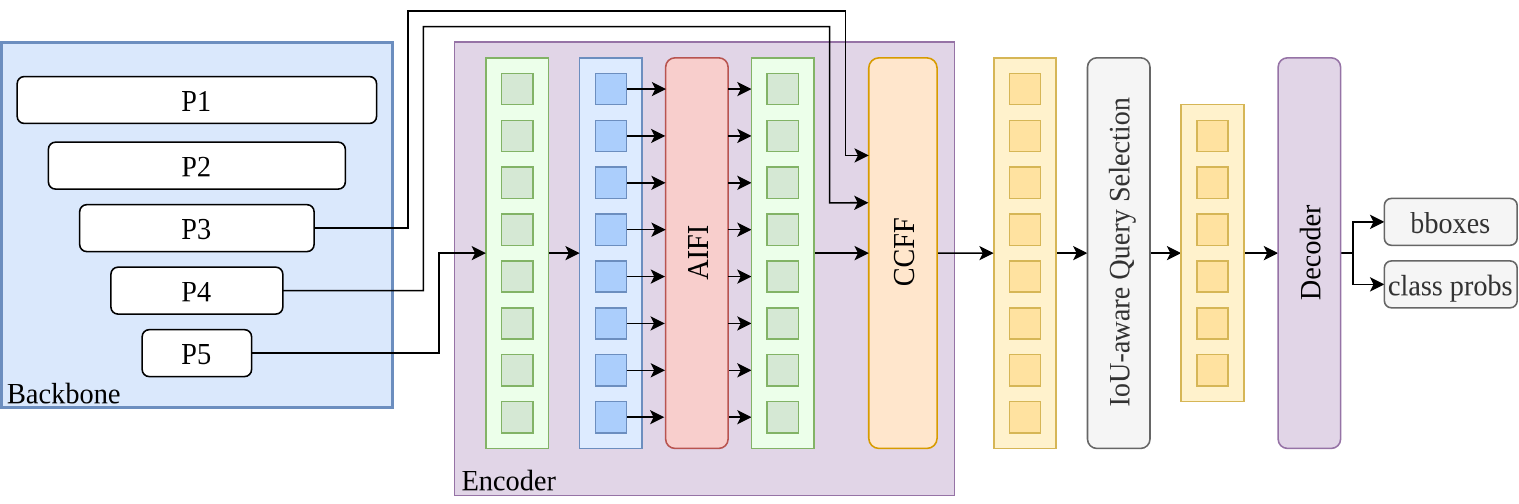
\caption{Diagram of RT-DETR architecture}
\label{fig:rtdetr}
\end{figure*}

As specimens occupy only a few pixels in the images, it is required to work with high-resolution images in order to retain valuable details that, otherwise, may get lost while applying an aggressive image resizing. In this scenario, modern and efficient models such as YOLO-based and DETR-based share a very favorable feature, their number of trainable weights parameters do not depend on the size of the image input. This is due to YOLO models being 'Fully Convolutional' networks and image size only determines the amount of times that kernel weights need to be applied. Although, this entails that increasing the image size does produce a higher computation demand and larger intermediate feature maps requiring more memory resources. Similarly, DETR models are able to work with an arbitrary image size without requiring to change the architecture, but the amount of internal features needed to be handled do increases with higher resolution images.
Taking into account these highly efficient capabilities of the models, it is decided to use their versions with the least amount of parameters along with 1920x1920 images. For YOLO, nano version YOLOv11n with 2.6 Million trainable parameters and the RT-DETR-l with approximately 45 Million parameters where selected. The RT-DETR-l is reported to provide real-time performance even with this high amount of parameters with images of 640x640 in \cite{zhao2024rtdetr} and is the smaller model with released weights to be fine-tuned, we considered interesting to evaluate its ability to scale with larger images. On the other hand, the nano versions of YOLOv11 represents one of the most efficient, proved and optimized models currently available.

An interesting aspect of the YOLO model is its ability to be specified to solve different tasks with the incorporation of specialized heads at the end of the architecture. We choose to experiment with two different variants of the YOLOv11, one with only the detection head that provides bounding boxes and classification label probabilities (termed as YOLOv11n-det), and a variant with an added instance segmentation head that provides precise masks of the detected specimens, delimiting its edges (YOLOv11n-seg). This type of experiment is inspired by He et al. \cite{he2017maskrcnn} where they show that adding a segmentation head improves also the performance of the detection task in the COCO dataset. The RT-DETR-l model do not provides a segmentation head and it is only trained to infer bounding boxes, classification label probabilities. 

\subsubsection{YOLOv11 Architecture}

This later variant is possible since our dataset provides precise segmentation masks and it is of interest to measure the impact of these masks. A schematic of the architecture, illustrating the flow of information through its various stages is shown in Figure \ref{fig:yolov11}. The network is typically structured into three primary components: the backbone, the neck, and the task-specific heads \cite{yolo_ultralytics}. The backbone is responsible for the initial extraction of visual features from the input image. It consists of convolutional layers organized into residual blocks with downsampling operations, progressively capturing semantic content at increasing levels of abstraction. The neck aggregates multi-scale features extracted by the backbone (Path Aggregation, PA), employing a faster variant of the Spatial Pyramid Pooling \cite{he2015spp} mechanism (SPPF block). Two variants of the Cross Stage Partial convolutional layers \cite{wang2020cspnet} are applied, the C3k2 block refers to a bottleneck version and the C2PSA block applies convolutional layers with Partial Spatial Attention (PSA) \cite{woo2018cbam}. The detection head, used in the YOLOv11n-det variant, predicts bounding boxes, class probabilities, and objectness scores at multiple scales. For the YOLOv11n-seg variant, an additional instance segmentation head is appended. This head predicts a set of prototype masks (proto-masks), shared across detections, along with mask coefficients for each object instance. The final segmentation mask for each object is generated as a linear combination of these proto-masks weighted by the corresponding coefficients \cite{bolya2019yolact}.

\subsubsection{RT-DETR Architecture}

The architecture of RT-DETR is composed of three major components: the backbone, the transformer encoder, and the transformer decoder \cite{zhao2024rtdetr}. The backbone operates similarly to that of YOLO-based models extracting feature maps from the input image using convolutional layers. The transformer encoder captures long-range dependencies in the image by attending to all spatial locations within the feature maps. This architecture enables RT-DETR to combine the global context modeling capabilities of transformers with the real-time performance of one-stage detectors. A visual summary of the RT-DETR model architecture is shown in Figure \ref{fig:rtdetr}. A key innovation in RT-DETR is the introduction of the Attention-based Intra-scale Feature Interaction (AIFI) module, that uses a single-scale transformer encoder applying the self-attention operation to only high-level features with richer semantic concepts (only P5 block). Afterwards, a CNN-based Cross-scale Feature Fusion block (CCFF) is in charge of integrate multi-scale features from the backbone (P3 and P4) with well-studied aggregation and pyramid pooling techniques similar to the ones present in the YOLO architecture \ref{fig:yolov11}. This mechanism enables efficient fusion of semantic and spatial information across scales, enhancing the robustness of the encoder to varying object sizes and positions. A novel feature selection module is in charge of selecting features with higher probability of being relevant (Query Selection) before passing them to a Transformer decoder which applies cross-attention to localize objects and generate contextual embeddings for classification and bounding box regression.

\begin{figure}[h!]
\centering
\def\svgwidth{width=0.9\linewidth}
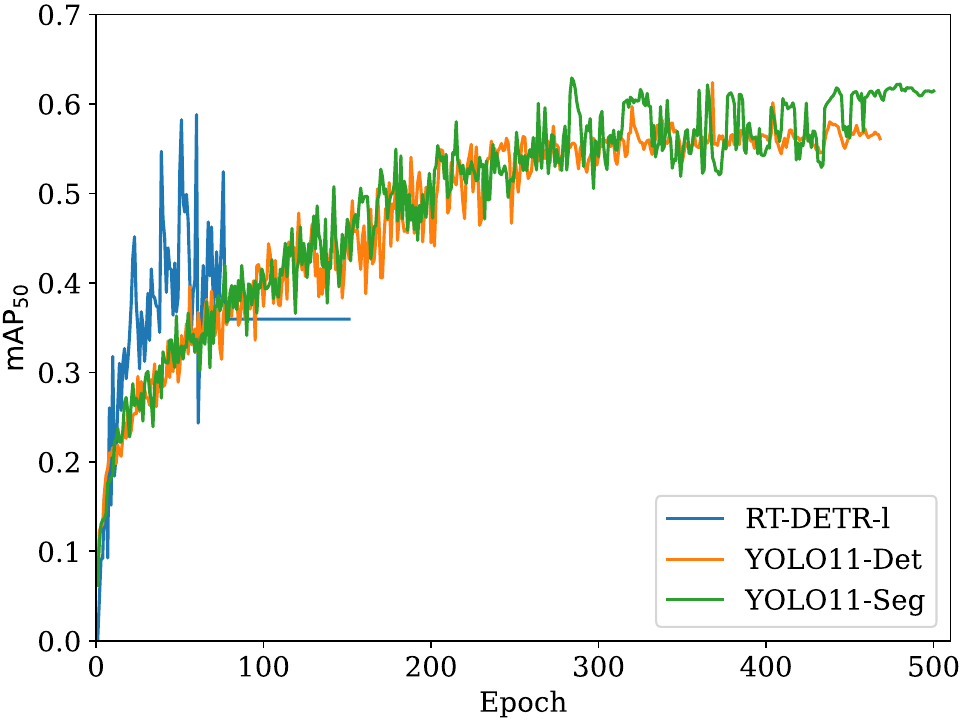
\caption{$mAP@50$ results in the validation set while training. RT-DETR-l model shows a faster convergence in less than 100 epochs where the $mAP@50$ stops increasing, while YOLO models show a softer convergence for around 450 epochs.}
\label{fig:training_map_metrics}
\end{figure}

\subsubsection{Training setup}

The model fine-tuning was performed using 160 images of deer with their respective masks, 275 images with cows, 138 images with capybaras, 41 images with birds,  10 images with horses, 3 images with axis deer, 6 images with people, 6 images with unclassified animals, and 575 images without any animals (one per transect). A validation set was designed containing 52 images of deer with their respective masks, 91 images with cows, 45 images with capybaras, 13 images with birds,  3 images with horses, 2 images with people, and 575 images without any animals (one per transect) in order to evaluate the models performance after every training epoch. The models where trained applying random image augmentation techniques such as image inversion, rotations and perturbations in the HSV color space. The used images were resized to 1920x1920, using a batch size of 8. The other parameters were maintained in the default value set by the Ultralytics repository \cite{yolo_ultralytics}.

\section{Experimentation}
\label{section:results}

Models were trained on an in-house cluster with a Nvidia T4 GPU node. The RT-DETR-l model took 6 hrs 8 min of training to complete 150 epochs, YOLOv11n-det took 6 hrs 41 min of training to complete 468 epochs and YOLOv11n-seg took 9 hrs 9 min of training to complete 500 epochs. The performance was evaluated using standard object detection metrics, Figure \ref{fig:training_map_metrics} shows the performance of the models in the validation set while being trained (mAP@50 with a fixed confidence threshold of 0.001), it can be seen that the YOLO models present a softer convergence with a consistent increase in the $mAP$, while the RT-DETR shows a very fast convergence in less than 75 epochs and the $mAP$ quickly stabilizes keeping a steady $mAP$ value. An early-stopping policy triggers in such cases where the RT-DETR model shows no signs of further improvement and training stops at 150 epochs, it is worth mentioning that the novel Transformer-based model introduced by the RT-DETR allows a very fast convergence finishing training in about the same time as YOLOv11n-det while having around 17 times more trainable parameters. The YOLOv11n-det model showed inference times between 250ms and 400ms, YOLOv11n-seg between 400ms and 500ms, and RT-DETR-l between 3100ms and 3200ms. It can be conclude that the RT-DETR architecture effectively loses its real-time capability as the image size increases.

Best model weights found during training stage where saved and used for further inspection of performance in the validation set.

\begin{figure}[h!]
\centering
\def\svgwidth{width=0.9\linewidth}
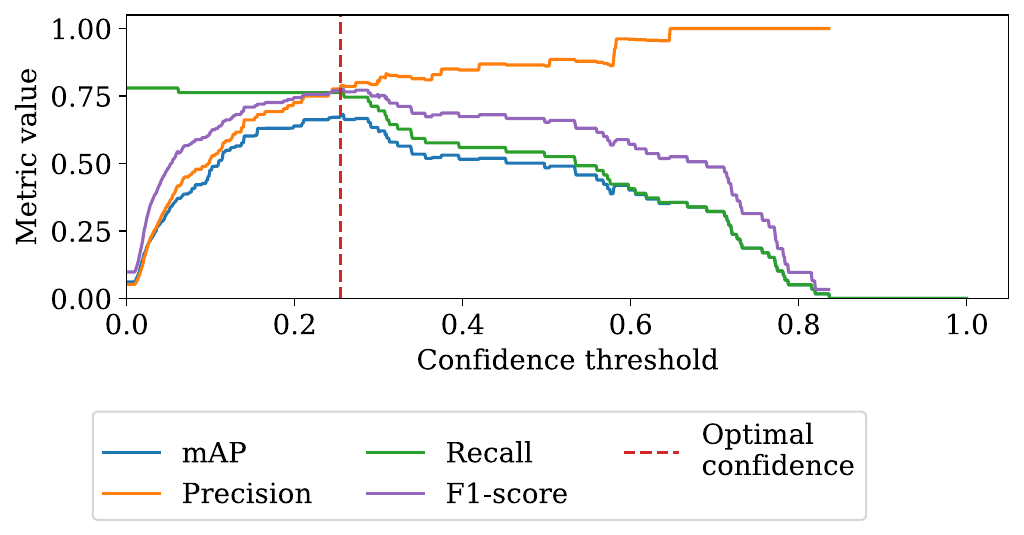
\caption{RT-DETR-l metrics obtained in the test set for different values of confidence threshold, with the established optimal confidence threshold 0.255 marked in red}
\label{fig:rtdetr_metrics}
\end{figure}

\begin{figure}[h!]
\centering
\def\svgwidth{width=0.9\linewidth}
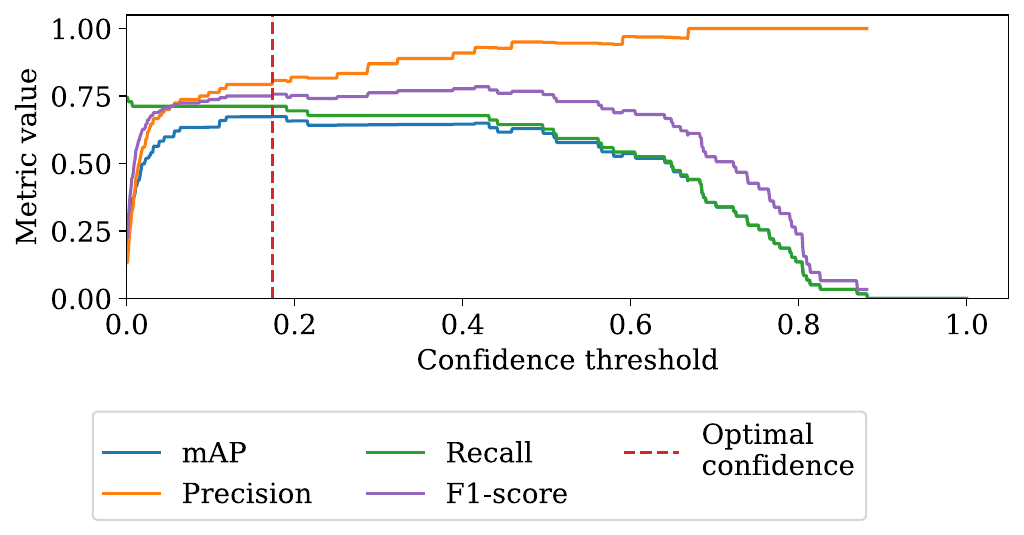
\caption{YOLOv11n-det metrics obtained in the test set for different values of confidence threshold, with the established optimal confidence threshold 0.174 marked in red}
\label{fig:yolo11ndet_metrics}
\end{figure}

\begin{figure}[h!]
\centering
\def\svgwidth{width=0.9\linewidth}
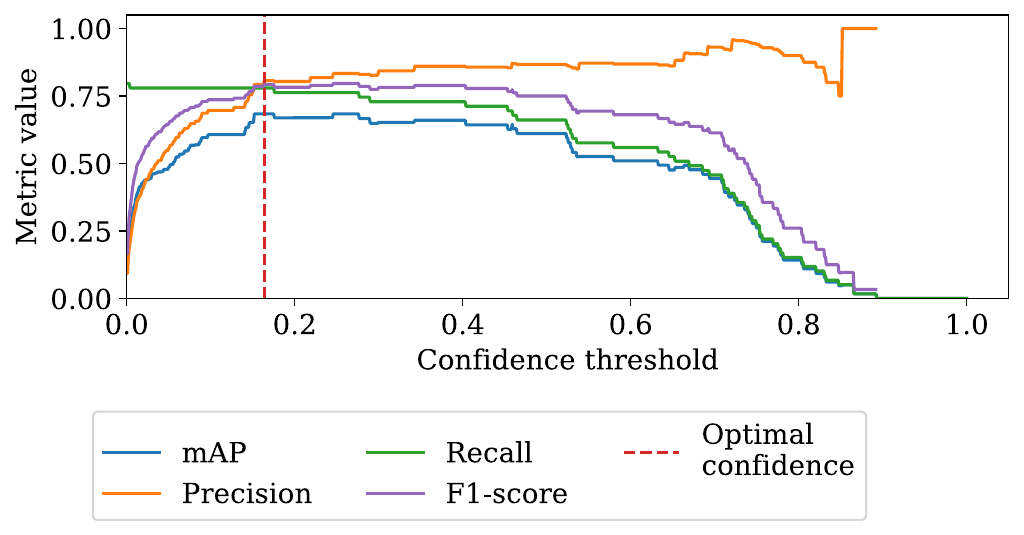
\caption{YOLOv11n-seg metrics obtained in the test set for different values of confidence threshold, with the established optimal confidence threshold 0.164 marked in red}
\label{fig:yolo11nseg_metrics}
\end{figure}

\subsection{Confidence Threshold and Overall Performance}

The optimal confidence threshold was selected based on the $mAP@50$ performance on the validation set. We used an IoU threshold of 50\% as it is a frequently used standard for comparing methods and the deer occupy a small area of the image and exact bounding box alignment is not crucial for this task. Figures \ref{fig:rtdetr_metrics}, \ref{fig:yolo11ndet_metrics} and \ref{fig:yolo11nseg_metrics} shows several metrics evaluated while inspecting different confidence thresholds, for each model an optimal confidence threshold was determined establishing a balance between precision, recall, the F1-score and the $mAP@50$. Optimal confidence thresholds 0.255, 0.174 and 0.164 where settle for RT-DETR-l, YOLOv11-det and YOLOv11-seg respectively.

The YOLOv11-seg was able to maintain a higher recall and F1-score that the YOLOv11-det, while the RT-DETR-l obtained very similar and competitive results with respect to the YOLOv11-seg even when does not take advantage of the segmentation masks.

\subsection{Classification Performance Evaluation}

\begin{figure}[h!]
\centering
\def\svgwidth{width=0.7\linewidth}
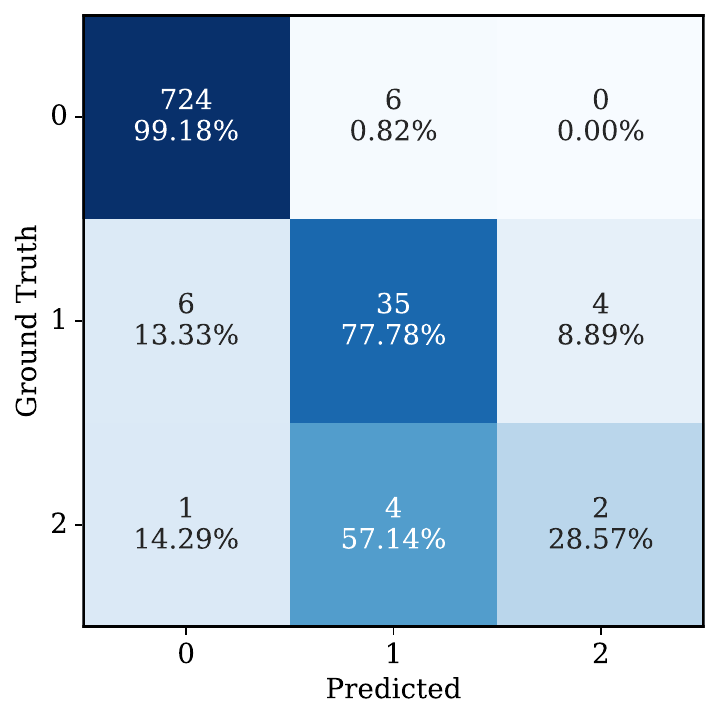
\caption{RT-DETR-l counting matrix using confidence threshold 0.255. Rows represent images where 0, 1 and 2 specimens are present, columns represent the predicted amount in each image.}
\label{fig:rtdetr_counting_matrix}
\end{figure}

\begin{figure}[h!]
\centering
\def\svgwidth{width=0.7\linewidth}
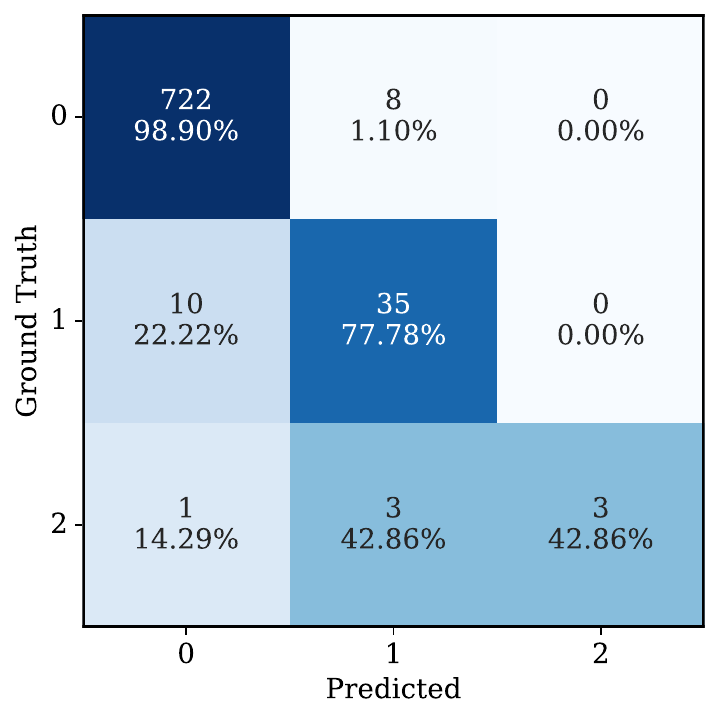
\caption{YOLOv11n-det counting matrix using confidence threshold 0.174. Rows represent images where 0, 1 and 2 specimens are present, columns represent the predicted amount in each image.}
\label{fig:yolo11ndet_counting_matrix}
\end{figure}

\begin{figure}[h!]
\centering
\def\svgwidth{width=0.7\linewidth}
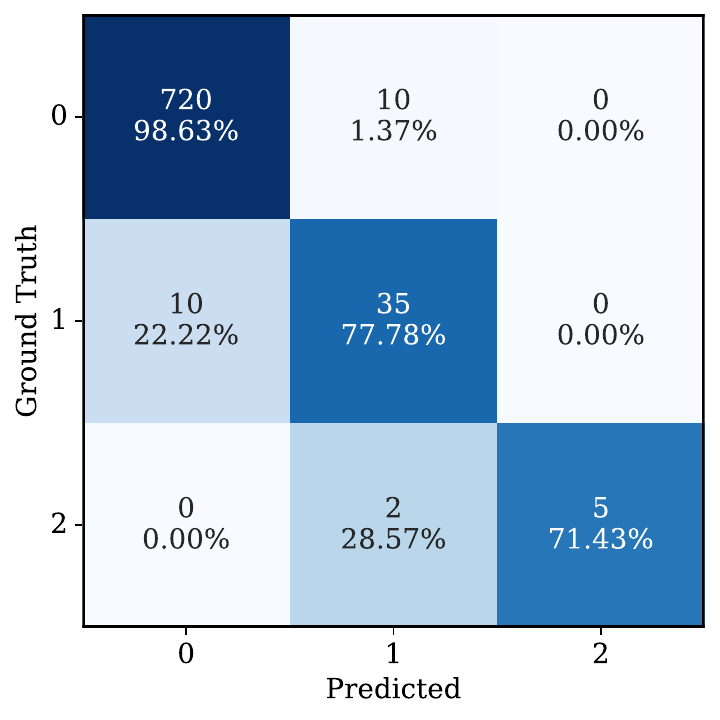
\caption{YOLOv11n-seg counting matrix using confidence threshold 0.164. Rows represent images where 0, 1 and 2 specimens are present, columns represent the predicted amount in each image.}
\label{fig:yolo11nseg_counting_matrix}
\end{figure}

In order to further assess the detection and classification labeling capabilities of the model, the number of detected specimens detected in each image was compared against the ground truth information. Each entry of the counting matrices shown in Figures \ref{fig:rtdetr_counting_matrix}, \ref{fig:yolo11ndet_counting_matrix} and \ref{fig:yolo11nseg_counting_matrix}, denote how many times the model predicted the correct (or incorrectly) amount of animals presents in the image. It can be seen that all models have a similar performance, being able to notice that the YOLOv11n-seg model was able to correctly discern cases where 2 animals where present in close proximity (diagonal bottom right value). A comparison of the detection of the different models in one of these cases can be seen in Figure \ref{fig:comparison}.

\subsection{Finding of Unlabeled Deer}

\begin{figure}[h!]
\centering
\includegraphics[width=0.8\linewidth]{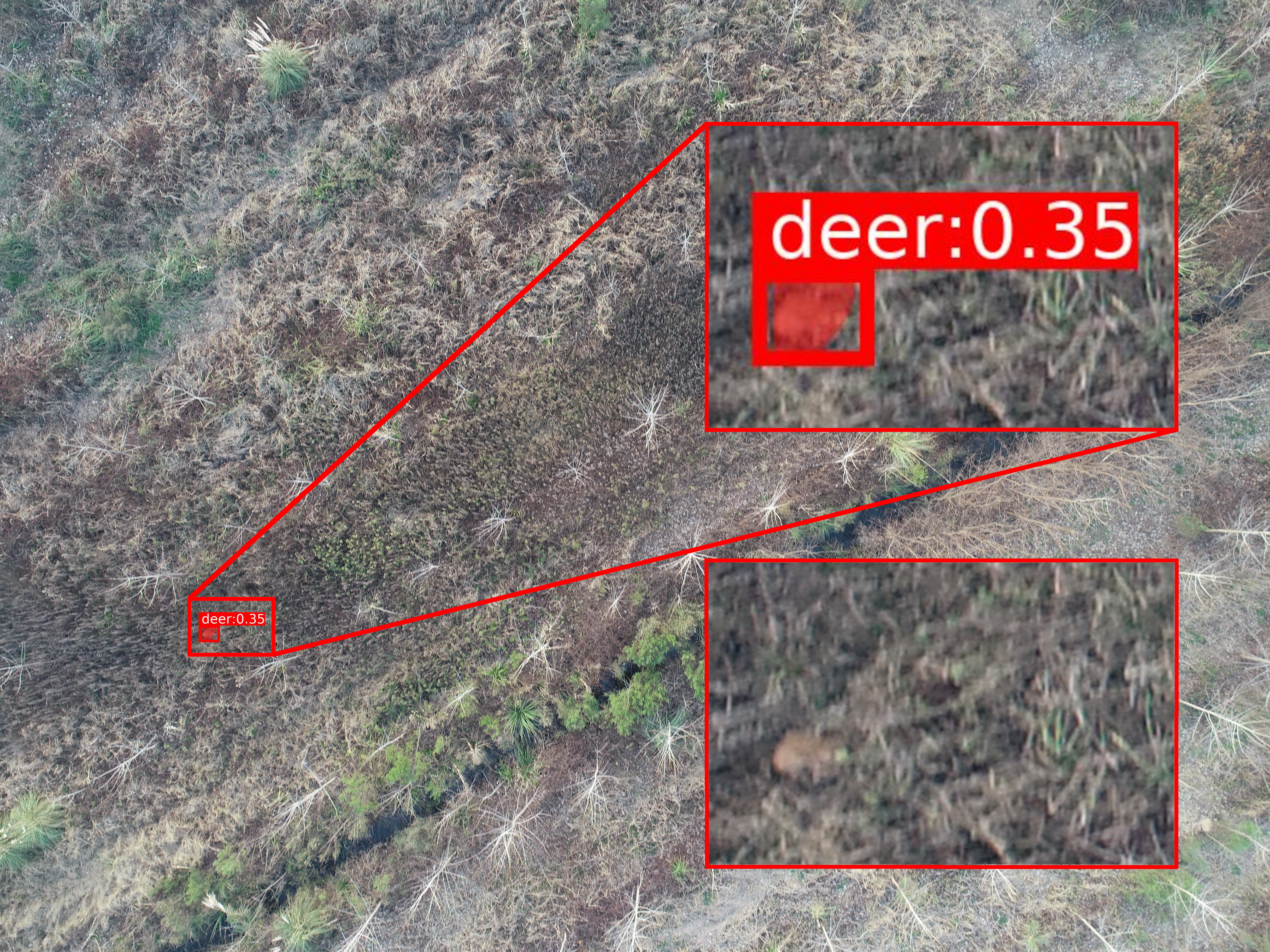}
\caption{Image with a segmented deer that had not originally been detected by the human labelers.}
\label{fig:lost_deer}
\end{figure}

In total, the model identified 54 deer that were initially missed, representing an increase of over 23\% in the total count. For instance, the deer shown in Fig.~\ref{fig:lost_deer} was discovered during the early iterations of the analysis. 

\FloatBarrier

\section{Conclusions and Future Work}
\label{section:conclusions}

These results lead us to the conclusion that the model YOLOv11n-seg is superior, showing greater potential in situations where there is overlapping specimens since it takes advantage of training considering the segmentation masks. The RT-DETR-l showed a slighly superior performance than the YOLOv11n-det model, achieving a very fast training convergence matching training times of much smaller models. Although, inference times of RT-DETR-l scale poorly with the image size being unable to work in real time with 1920x1920 images, it would be worth to explore smaller variants of the RT-DETR model in the future.


\bibliographystyle{IEEEtran}
\bibliography{referencias}

\end{document}

%% file: img/yolo11.drawio.pdf_tex
\begingroup%
  \makeatletter%
  \providecommand\color[2][]{%
    \errmessage{(Inkscape) Color is used for the text in Inkscape, but the package 'color.sty' is not loaded}%
    \renewcommand\color[2][]{}%
  }%
  \providecommand\transparent[1]{%
    \errmessage{(Inkscape) Transparency is used (non-zero) for the text in Inkscape, but the package 'transparent.sty' is not loaded}%
    \renewcommand\transparent[1]{}%
  }%
  \providecommand\rotatebox[2]{#2}%
  \newcommand*\fsize{\dimexpr\f@size pt\relax}%
  \newcommand*\lineheight[1]{\fontsize{\fsize}{#1\fsize}\selectfont}%
  \ifx\svgwidth\undefined%
    \setlength{\unitlength}{564.75bp}%
    \ifx\svgscale\undefined%
      \relax%
    \else%
      \setlength{\unitlength}{\unitlength * \real{\svgscale}}%
    \fi%
  \else%
    \setlength{\unitlength}{\linewidth}%
  \fi%
  \global\let\svgwidth\undefined%
  \global\let\svgscale\undefined%
  \makeatother%
  \begin{picture}(1,0.50066401)%
    \lineheight{1}%
    \setlength\tabcolsep{0pt}%
    \put(0,0){\includegraphics[width=\unitlength,page=1]{img/yolo11.drawio.pdf}}%
  \end{picture}%
\endgroup%

%% file: img/rtdetr.drawio.pdf_tex
\begingroup%
  \makeatletter%
  \providecommand\color[2][]{%
    \errmessage{(Inkscape) Color is used for the text in Inkscape, but the package 'color.sty' is not loaded}%
    \renewcommand\color[2][]{}%
  }%
  \providecommand\transparent[1]{%
    \errmessage{(Inkscape) Transparency is used (non-zero) for the text in Inkscape, but the package 'transparent.sty' is not loaded}%
    \renewcommand\transparent[1]{}%
  }%
  \providecommand\rotatebox[2]{#2}%
  \newcommand*\fsize{\dimexpr\f@size pt\relax}%
  \newcommand*\lineheight[1]{\fontsize{\fsize}{#1\fsize}\selectfont}%
  \ifx\svgwidth\undefined%
    \setlength{\unitlength}{729bp}%
    \ifx\svgscale\undefined%
      \relax%
    \else%
      \setlength{\unitlength}{\unitlength * \real{\svgscale}}%
    \fi%
  \else%
    \setlength{\unitlength}{\linewidth}%
  \fi%
  \global\let\svgwidth\undefined%
  \global\let\svgscale\undefined%
  \makeatother%
  \begin{picture}(1,0.33179012)%
    \lineheight{1}%
    \setlength\tabcolsep{0pt}%
    \put(0,0){\includegraphics[width=\unitlength,page=1]{img/rtdetr.drawio.pdf}}%
  \end{picture}%
\endgroup%

%% file: img/results/training_mAP50.pdf_tex
\begingroup%
  \makeatletter%
  \providecommand\color[2][]{%
    \errmessage{(Inkscape) Color is used for the text in Inkscape, but the package 'color.sty' is not loaded}%
    \renewcommand\color[2][]{}%
  }%
  \providecommand\transparent[1]{%
    \errmessage{(Inkscape) Transparency is used (non-zero) for the text in Inkscape, but the package 'transparent.sty' is not loaded}%
    \renewcommand\transparent[1]{}%
  }%
  \providecommand\rotatebox[2]{#2}%
  \newcommand*\fsize{\dimexpr\f@size pt\relax}%
  \newcommand*\lineheight[1]{\fontsize{\fsize}{#1\fsize}\selectfont}%
  \ifx\svgwidth\undefined%
    \setlength{\unitlength}{460.8bp}%
    \ifx\svgscale\undefined%
      \relax%
    \else%
      \setlength{\unitlength}{\unitlength * \real{\svgscale}}%
    \fi%
  \else%
    \setlength{\unitlength}{\linewidth}%
  \fi%
  \global\let\svgwidth\undefined%
  \global\let\svgscale\undefined%
  \makeatother%
  \begin{picture}(1,0.75)%
    \lineheight{1}%
    \setlength\tabcolsep{0pt}%
    \put(0,0){\includegraphics[width=\unitlength,page=1]{img/results/training_mAP50.pdf}}%
  \end{picture}%
\endgroup%

%% file: img/results/confidence/rtdetr/rtdetr_metrics.pdf_tex
\begingroup%
  \makeatletter%
  \providecommand\color[2][]{%
    \errmessage{(Inkscape) Color is used for the text in Inkscape, but the package 'color.sty' is not loaded}%
    \renewcommand\color[2][]{}%
  }%
  \providecommand\transparent[1]{%
    \errmessage{(Inkscape) Transparency is used (non-zero) for the text in Inkscape, but the package 'transparent.sty' is not loaded}%
    \renewcommand\transparent[1]{}%
  }%
  \providecommand\rotatebox[2]{#2}%
  \newcommand*\fsize{\dimexpr\f@size pt\relax}%
  \newcommand*\lineheight[1]{\fontsize{\fsize}{#1\fsize}\selectfont}%
  \ifx\svgwidth\undefined%
    \setlength{\unitlength}{491.351667bp}%
    \ifx\svgscale\undefined%
      \relax%
    \else%
      \setlength{\unitlength}{\unitlength * \real{\svgscale}}%
    \fi%
  \else%
    \setlength{\unitlength}{\linewidth}%
  \fi%
  \global\let\svgwidth\undefined%
  \global\let\svgscale\undefined%
  \makeatother%
  \begin{picture}(1,0.5225748)%
    \lineheight{1}%
    \setlength\tabcolsep{0pt}%
    \put(0,0){\includegraphics[width=\unitlength,page=1]{img/results/confidence/rtdetr/rtdetr_metrics.pdf}}%
  \end{picture}%
\endgroup%

%% file: img/results/confidence/yolo11-det/det_metrics.pdf_tex
\begingroup%
  \makeatletter%
  \providecommand\color[2][]{%
    \errmessage{(Inkscape) Color is used for the text in Inkscape, but the package 'color.sty' is not loaded}%
    \renewcommand\color[2][]{}%
  }%
  \providecommand\transparent[1]{%
    \errmessage{(Inkscape) Transparency is used (non-zero) for the text in Inkscape, but the package 'transparent.sty' is not loaded}%
    \renewcommand\transparent[1]{}%
  }%
  \providecommand\rotatebox[2]{#2}%
  \newcommand*\fsize{\dimexpr\f@size pt\relax}%
  \newcommand*\lineheight[1]{\fontsize{\fsize}{#1\fsize}\selectfont}%
  \ifx\svgwidth\undefined%
    \setlength{\unitlength}{491.351667bp}%
    \ifx\svgscale\undefined%
      \relax%
    \else%
      \setlength{\unitlength}{\unitlength * \real{\svgscale}}%
    \fi%
  \else%
    \setlength{\unitlength}{\linewidth}%
  \fi%
  \global\let\svgwidth\undefined%
  \global\let\svgscale\undefined%
  \makeatother%
  \begin{picture}(1,0.5225748)%
    \lineheight{1}%
    \setlength\tabcolsep{0pt}%
    \put(0,0){\includegraphics[width=\unitlength,page=1]{img/results/confidence/yolo11-det/det_metrics.pdf}}%
  \end{picture}%
\endgroup%

%% file: img/results/confidence/yolo11-seg/seg_metrics.pdf_tex
\begingroup%
  \makeatletter%
  \providecommand\color[2][]{%
    \errmessage{(Inkscape) Color is used for the text in Inkscape, but the package 'color.sty' is not loaded}%
    \renewcommand\color[2][]{}%
  }%
  \providecommand\transparent[1]{%
    \errmessage{(Inkscape) Transparency is used (non-zero) for the text in Inkscape, but the package 'transparent.sty' is not loaded}%
    \renewcommand\transparent[1]{}%
  }%
  \providecommand\rotatebox[2]{#2}%
  \newcommand*\fsize{\dimexpr\f@size pt\relax}%
  \newcommand*\lineheight[1]{\fontsize{\fsize}{#1\fsize}\selectfont}%
  \ifx\svgwidth\undefined%
    \setlength{\unitlength}{491.351667bp}%
    \ifx\svgscale\undefined%
      \relax%
    \else%
      \setlength{\unitlength}{\unitlength * \real{\svgscale}}%
    \fi%
  \else%
    \setlength{\unitlength}{\linewidth}%
  \fi%
  \global\let\svgwidth\undefined%
  \global\let\svgscale\undefined%
  \makeatother%
  \begin{picture}(1,0.5225748)%
    \lineheight{1}%
    \setlength\tabcolsep{0pt}%
    \put(0,0){\includegraphics[width=\unitlength,page=1]{img/results/confidence/yolo11-seg/seg_metrics.pdf}}%
  \end{picture}%
\endgroup%

%% file: img/results/confidence/rtdetr/rtdetr_confusion_matrix.pdf_tex
\begingroup%
  \makeatletter%
  \providecommand\color[2][]{%
    \errmessage{(Inkscape) Color is used for the text in Inkscape, but the package 'color.sty' is not loaded}%
    \renewcommand\color[2][]{}%
  }%
  \providecommand\transparent[1]{%
    \errmessage{(Inkscape) Transparency is used (non-zero) for the text in Inkscape, but the package 'transparent.sty' is not loaded}%
    \renewcommand\transparent[1]{}%
  }%
  \providecommand\rotatebox[2]{#2}%
  \newcommand*\fsize{\dimexpr\f@size pt\relax}%
  \newcommand*\lineheight[1]{\fontsize{\fsize}{#1\fsize}\selectfont}%
  \ifx\svgwidth\undefined%
    \setlength{\unitlength}{345.365bp}%
    \ifx\svgscale\undefined%
      \relax%
    \else%
      \setlength{\unitlength}{\unitlength * \real{\svgscale}}%
    \fi%
  \else%
    \setlength{\unitlength}{\linewidth}%
  \fi%
  \global\let\svgwidth\undefined%
  \global\let\svgscale\undefined%
  \makeatother%
  \begin{picture}(1,0.99931473)%
    \lineheight{1}%
    \setlength\tabcolsep{0pt}%
    \put(0,0){\includegraphics[width=\unitlength,page=1]{img/results/confidence/rtdetr/rtdetr_confusion_matrix.pdf}}%
  \end{picture}%
\endgroup%

%% file: img/results/confidence/yolo11-det/det_confusion_matrix.pdf_tex
\begingroup%
  \makeatletter%
  \providecommand\color[2][]{%
    \errmessage{(Inkscape) Color is used for the text in Inkscape, but the package 'color.sty' is not loaded}%
    \renewcommand\color[2][]{}%
  }%
  \providecommand\transparent[1]{%
    \errmessage{(Inkscape) Transparency is used (non-zero) for the text in Inkscape, but the package 'transparent.sty' is not loaded}%
    \renewcommand\transparent[1]{}%
  }%
  \providecommand\rotatebox[2]{#2}%
  \newcommand*\fsize{\dimexpr\f@size pt\relax}%
  \newcommand*\lineheight[1]{\fontsize{\fsize}{#1\fsize}\selectfont}%
  \ifx\svgwidth\undefined%
    \setlength{\unitlength}{345.365bp}%
    \ifx\svgscale\undefined%
      \relax%
    \else%
      \setlength{\unitlength}{\unitlength * \real{\svgscale}}%
    \fi%
  \else%
    \setlength{\unitlength}{\linewidth}%
  \fi%
  \global\let\svgwidth\undefined%
  \global\let\svgscale\undefined%
  \makeatother%
  \begin{picture}(1,0.99931473)%
    \lineheight{1}%
    \setlength\tabcolsep{0pt}%
    \put(0,0){\includegraphics[width=\unitlength,page=1]{img/results/confidence/yolo11-det/det_confusion_matrix.pdf}}%
  \end{picture}%
\endgroup%

%% file: img/results/confidence/yolo11-seg/seg_confusion_matrix.pdf_tex
\begingroup%
  \makeatletter%
  \providecommand\color[2][]{%
    \errmessage{(Inkscape) Color is used for the text in Inkscape, but the package 'color.sty' is not loaded}%
    \renewcommand\color[2][]{}%
  }%
  \providecommand\transparent[1]{%
    \errmessage{(Inkscape) Transparency is used (non-zero) for the text in Inkscape, but the package 'transparent.sty' is not loaded}%
    \renewcommand\transparent[1]{}%
  }%
  \providecommand\rotatebox[2]{#2}%
  \newcommand*\fsize{\dimexpr\f@size pt\relax}%
  \newcommand*\lineheight[1]{\fontsize{\fsize}{#1\fsize}\selectfont}%
  \ifx\svgwidth\undefined%
    \setlength{\unitlength}{345.365bp}%
    \ifx\svgscale\undefined%
      \relax%
    \else%
      \setlength{\unitlength}{\unitlength * \real{\svgscale}}%
    \fi%
  \else%
    \setlength{\unitlength}{\linewidth}%
  \fi%
  \global\let\svgwidth\undefined%
  \global\let\svgscale\undefined%
  \makeatother%
  \begin{picture}(1,0.99931473)%
    \lineheight{1}%
    \setlength\tabcolsep{0pt}%
    \put(0,0){\includegraphics[width=\unitlength,page=1]{img/results/confidence/yolo11-seg/seg_confusion_matrix.pdf}}%
  \end{picture}%
\endgroup%